%% file: 01CAH_emnlp.tex
\pdfoutput=1

\documentclass[11pt]{article}
\usepackage{graphicx}
\usepackage{svg}
\usepackage{amsmath}
\usepackage[scientific-notation=true]{siunitx}
\usepackage{paralist}
\usepackage{EMNLP2022}
\usepackage{xcolor}
\usepackage{times}
\usepackage{latexsym}
\usepackage{caption}
\usepackage{subcaption}
\usepackage[T1]{fontenc}
\usepackage{natbib}
\usepackage{parskip}

\usepackage[utf8]{inputenc}

\usepackage{microtype}

\usepackage{inconsolata}

\usepackage{color}
\newcommand{\dnote}[1]{\textcolor{red}{$\ll$\textsf{#1 | Dafna}$\gg$}}
\newcommand{\donote}[1]{\textcolor{green}{$\ll$\textsf{#1 | Dan}$\gg$}}

\newcommand{\remove}[1]{}
\newcommand{\xhdr}[1]{\vspace{1mm}\noindent{{\bf #1.}}} 
\title{Cards Against AI: Predicting Humor in a Fill-in-the-blank Party Game}

\author{Dan Ofer \\
 The Hebrew University of Jerusalem \\
  \texttt{dan.ofer@mail.huji.ac.il} \\\And
  Dafna Shahaf \\
  The Hebrew University of Jerusalem \\
  \texttt{dshahaf@cs.huji.ac.il} \\
}

\begin{document}
\maketitle


\begin{abstract}
Humor is an inherently social phenomenon, with humorous utterances shaped by what is socially and culturally accepted. 
Understanding humor is an important NLP challenge, with many applications to human-computer interactions. In this work we explore humor in the context of Cards Against Humanity -- a party game where players complete fill-in-the-blank statements using cards that can be offensive or politically incorrect.
We introduce a novel dataset of 300,000 online games of Cards Against Humanity, including 785K unique jokes, analyze it and provide insights. We trained machine learning models to predict the winning joke per game, achieving performance twice as good (20\%) as random, even without any user information.
On the more difficult task of judging novel cards, we see the models' ability to generalize is moderate. 
Interestingly, we find that our models are primarily focused on punchline card, with the context having little impact.
Analyzing feature importance, we observe that short, crude, juvenile punchlines tend to win.
\end{abstract}

\section{Introduction}

\input{01alternativeintro}

\section{Data}
The dataset consists of games played on the online CAH labs website, \url{https://lab.cardsagainsthumanity.com}. The players played the game voluntarily, for
fun; they are not our annotators or workers. In each round a user is presented with a random  prompt card, 10 potential punchlines cards, and chooses the funniest punchline. The raw data had 298,955 past games (i.e., we did not perform any additional experimentation ourselves).
There are 581 unique black prompt cards and 2,128 white punchline cards, including cards from the official CAH game and expansions, resulting in 1,236,368 possible unique \emph{jokes} (where a joke is the result of filling in the blank of the prompt card with a punchline). {Each round is effectively unique due to the large number of combinations.}  
%
The data we received from CAH did not include any demographic or geographic characteristics, user identifiers or personally identifiable information. 
5\% of games were skipped by users and were excluded, as were a minority of prompts that required picking more than one punchline. 
Data is available upon request to CAH at \url{mail@cardsagainsthumanity.com}. 

\subsection{Data analysis and observations}\label{sec:analysis}
\begin{figure}[t!]
\includegraphics[width=0.96\linewidth]{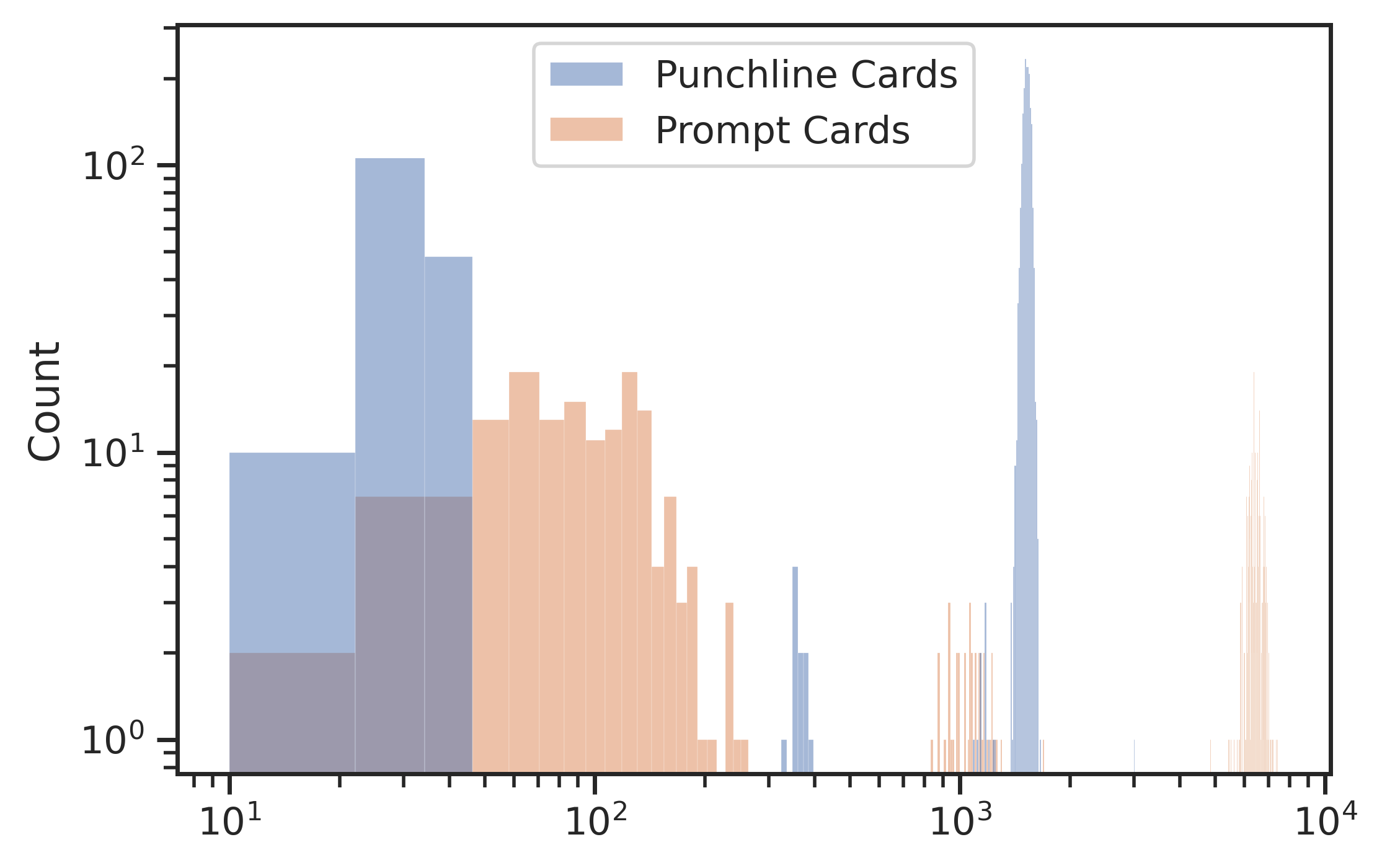}
\caption{Card counts. Log scale histogram of prompt and punchline cards occurrence frequency (i.e., how many times cards appeared). 
Prompts have a more relatively uniform distribution, but both prompts and punchline cards have a ``tail'' of rare cards. The spikes of frequent cards are presumably due to cards from the standard game, as opposed to experimental cards or expansions.}
\label{fig:1a}
\end{figure}

\begin{figure}[t!]
\includegraphics[width=0.96\linewidth]{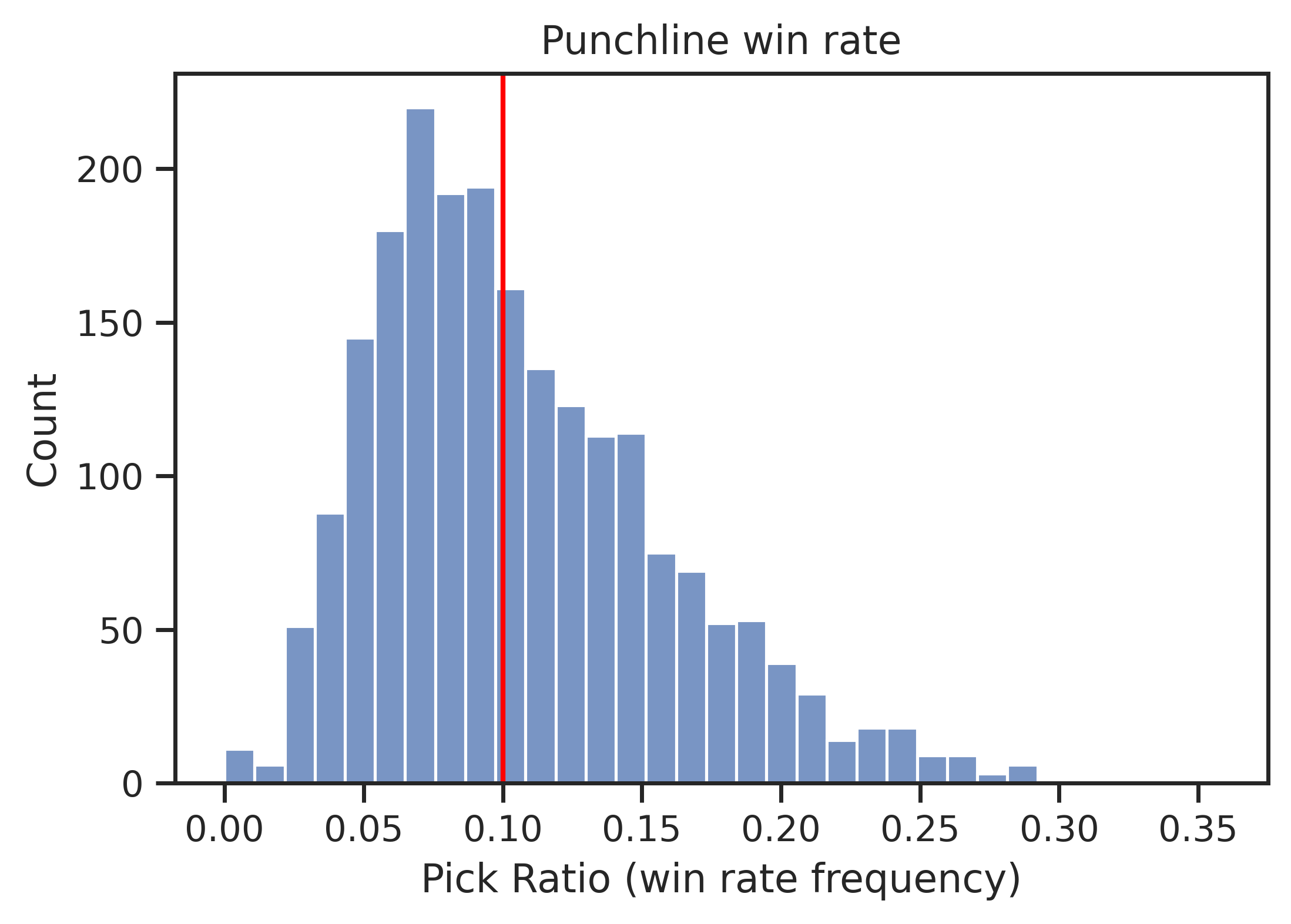}
\caption{Punchline win rate frequency (Pick ratio) of punchlines, across all games. Random is 0.1}
\label{fig:1b}
\end{figure}

The frequency of different prompts or punchlines presented to users is not a uniform distribution (Fig \ref{fig:1a}). The odds of a punchline card being picked and winning is also unevenly distributed -- perhaps unsurprisingly, some punchlines are funnier than others (Fig \ref{fig:1b}).
The data is sparse: the number of potential games is immense (\(7.06 \times 10^{54})\). Viewed at the level of unique jokes (prompt+punchline combined), only 784,974 appear at least once across the games, out of the 1.23M possible (60\%), with few repeats. If we consider only cases where we have feedback (a ``winning pick''), then we have only 248,896 jokes with feedback, and of these 77\% were picked only once, out of ~300,000 games. A further 17\% were picked only twice. 



\subsubsection{Popular punchlines}\label{subsection:punchline analysis}
Across all games, all punchlines appeared at least 14 times, with $\mu =1149$, $\sigma =334$. 
We considered a punchline successful if its win rate is over 20\% (twice better than random). Dirty, short and explicit punchlines dominated the list of successful punchlines. 
(Censored) punchlines include 
{\it Syphilis, Incest, 
The death penalty, 
COVID-19,} and {\it Joe Biden.}

Unsuccessful punchlines (win rate of under 2\%, 5 times worse than random) include 
{\it Being seen reading Infinite Jest, Usury, Running afoul of the sultans Janissaries, The significance of eyes in King Lear,} and a card Neil Gaiman famously wrote: {\it Three elves at a time.} Many unsuccessful cards are long or contain obscure references. 
All examples have p-value < \num{1e-25} (two sided binomial test). 

\subsubsection{Funny combinations}
We set out to find successful combinations, meaning jokes that outperformed the baseline of their constituent punchline.  Jokes were ranked using a two-sided binomial test. A success was defined as the number of times a joke was picked, the number of events as the number of times the joke occurred, and prior success rate defined as the prior win rate of the punchline. In other words, we compare the win rate of a joke to that expected from just the presence of the punchline.
%
%
Some significantly (\emph{p}<\num{1e-4}) overperforming (censored) jokes include: 

\begin{compactitem}
\item What makes life worth living? \underline{Dying}
\item 	She's up all night for good fun. I'm up all night for \underline{like 50 mosquito bites}.
\item 	The Japanese have developed a smaller, more efficient version of \underline{emotional unavailability}.
\end{compactitem}

There were very few significantly underperforming pairs, and all involved overly crass examples, so we do not include them here.

\section{Predicting winning jokes}\label{task1}
\input{03task1}

\section{Novel punchlines}\label{task2}
\input{04task2}


\section{Discussion and Conclusions}
In this work we explore humor in the context of the popular card game Cards Against Humanity.
We introduce a novel dataset of 300,000 online games and 785K unique jokes, analyze it and provide insights. We trained state-of-the-art machine learning models to predict the winning joke per game. Interestingly, we find that past performance of the punchline card is a very strong indicator (unrelated to the prompt), and that short and crude punchlines tend to win. We show our models primarily focus on punchline, and observe potential  {behavioral biases} in the data. 
On the more difficult task of judging novel cards, we see the models' ability to generalize is moderate, leaving room for future work. 
We believe humor is a crucial component in developing
personable human-computer interactions, and the CAH dataset has several characteristics rendering
it particularly attractive for NLP research. We hope it will promote further work in this area.

\section{Limitations and bias}
The population who play the game online may not be representative of the overall population, or of CAH players in general. As such, the results should not be taken as a definitive guide to what types of jokes are most likely to be found funny by the general population.
As we only had access to limited data, the study did not consider some potentially important factors such as user demographics, context, or the interaction between players.

We note that humor is highly subjective and dependent on context and user characteristics.
However, we believe that there is still value in the data we have: studies have shown some humor techniques are consistently funny across different cultures (e.g., exaggeration, understatement, disguise, deception, and resolution of incongruities). In other words, there are some universally appreciated kinds of humor. We also note that many of the cards are culture-specific (e.g., Judge Judy and Walmart) and we carefully posit that certain demographics are more likely to play the game in the first place.

\section*{Ethics Statement}
The data we used in this work contains extreme speech
that can be shocking and offensive, especially to protected/sensitive/minority groups. 
The jokes and examples do not represent our opinions, or those of anyone involved. This data is not intended for and should not be used for training
models applied to real-world tasks, since such models might exhibit and
propagate those offensive messages. We do believe that it is important to study offensive humor as a way to understand its role in generating and reinforcing social boundaries and inequalities. The authors are unaffiliated with CAH, and have no competing interests.

All our data came from the CAH website, where players had played the game voluntarily, for fun. They are not workers and not pressured to participate. We received access to past games, i.e., we did not perform any additional experimentation ourselves. 
CAH are sensitive to removing personally identifiable information; their privacy policy (\url{https://www.cardsagainsthumanity.com/privacy-policy}) clearly states what data is gathered and limitations on third party disclosure. The dataset contained only round and card-level data (e.g., card texts, round duration). All user identifiable data, including any demographic or geographic characteristics, was removed before we accessed it.
The study received IRB approval from the Hebrew University of Jerusalem.

\paragraph{Acknowledgements}
{We thank the anonymous reviewers for their constructive comments and Cards Against Humanity for the data. This work was supported by the European Research Council (ERC) under the European Union’s Horizon
2020 research and innovation programme (grant
no. 852686, SIAM).}

 \bibliography{anthology,custom,MyLibrary,cites2}
 \bibliographystyle{acl_natbib}




\end{document}

%% file: 01alternativeintro.tex
Humor is a universal phenomenon, fulfilling important social roles: approaching social taboos, expressing criticism against individuals and institutions, and consolidating a sense of belonging to a group \cite{ziv2010social}. Humorous utterances are shaped by what is socially and culturally accepted. 

Humor underpins many social interactions  \cite{beach2017laughter,urbatsch2022humor}. It increases likeability and trust \cite{meyer2015understanding}. Thus, humor is also a crucial component in developing personable human-computer interactions. 


Specifically, we focus on the task of \emph{humor recognition} -- determining whether a sentence in a given context is funny. This task is difficult, as
humor is a diverse, amorphous and complex phenomenon. It
requires creativity and common sense, and is very challenging to model
\cite{Winters2021Computers,attardo2010linguistic},
 considered by some researchers to be AI-complete \cite{stock2003hahacronym}. Thus, 
designing a general humor recognition algorithm currently seems beyond our reach, and works on computational humor tend to focus on narrow, specific types of humor, such as knock-knock jokes, one-liners, or even that's-what-she-said jokes \cite{https://doi.org/10.1111/j.1467-8640.2006.00278.x,taylor_computationally_2004,kiddon-brun-2011-thats} 

In this work we explore humor in the context of the immensely popular card game {\bf ``Cards Against Humanity” (CAH)}. 
The game mechanics are simple: 
Players are dealt ten cards (``punchlines''). The judge of the round draws a ``prompt'' card posing a question or a ``fill-in-the-blank'' statement. Each player submits an answer from their hand, and the judge picks the winner.
An example prompt is ``TSA guidelines now prohibit \_\_\_ on airplanes''. Candidate punchlines are ``Goblins'', ``BATMAN!!!'',  ``Poor people'', and ``The right amount of cocaine''. Importantly, many cards are offensive or politically incorrect.

We introduce a novel dataset of 300K online CAH games. While most current humor datasets are lacking in size \cite{weller-seppi-2020-rjokes}, or have weak labels (e.g., upvotes without total views), our dataset is large and strongly labeled. We train machine learning models\footnote{Code  available at   \url{https://github.com/ddofer/CAH}} 
to predict the winning joke per round and show models can somewhat generalize to novel (unseen) punchline cards. Surprisingly, we find that our models primarily focus on the \emph{punchline card alone}, and the impact of the prompt  is limited. We also identify potential behavioral biases in 
the data.

Our main goal here is to explore humor through a data-driven lens, and we believe CAH provides a unique perspective to this end.
Most existing studies on humor recognition 
formulate the problem as a binary classification task and
try to recognize jokes via a set of linguistic
features \cite{yang2015humor,purandare2006humor,zhang2014recognizing}. One of the common problems those works face is the construction of negative instances, which are often sampled from a different domain (e.g., news). In contrast, the CAH task does not suffer from this problem.

Perhaps the closest setting to ours is humorous fill-in-the-blank  \cite{hossain2017filling,https://doi.org/10.48550/arxiv.2006.00578}, where users complete a joke however they see fit. However, our setting is a lot more restricted: players choose  (\emph{rank}) an answer from a small set of options,
%
enabling \emph{comparisons} that would be hard to test on other corpora.

From a humor-theory point of view, we believe CAH serves as an interesting example of \emph{frame blends} and \emph{frame shifts} \cite{hofstadter1989synopsis,coulson_2001}, where a speaker's mental model suddenly shifts to new situations, or two distinct situations create a hybrid.  
CAH provides a
relatively clean setting to explore this phenomenon, as the jokes are short, with simple syntax and narrative structure.



To the best of our knowledge, CAH has only been explored in the literature through pedagogical, ethical or sociological lenses (e.g., \cite{strmic2016equal}), not computational or linguistic ones.
We note the data contains offensive humor, and should be very carefully used as training data. However, we believe it is  important to study offensive humor too and understand its role in generating and reinforcing social boundaries and inequalities.

%% file: 03task1.tex

Given a prompt and 10 punchlines (i.e., a round), our goal in this section is to predict the punchline most likely to be picked as funniest.
We note that unlike in traditional user-item ranking or recommendation tasks, we lack any user information.

\subsection{Methods}\label{methods1}

 The data was split at the level of rounds into train and test subsets (80/20 split), with 195,708 rounds (746K unique jokes) for training, and 48,928 rounds (367K unique jokes) as the test set. {The IDs are provided in the repository for the sake of reproducability.} Target distribution was identical (10\%). We tried the following methods:

 \xhdr{Joke popularity baseline} 
 A simple baseline computing the prior mean win frequency of the combination (joke) in the train set. Jokes not in train were imputed with the mean. We experimented with global smoothing and minimum occurrences, but found that this did not improve the baseline.

    \xhdr{Punchline popularity baseline}  
 Similar to joke popularity, but using prior win frequency of the punchline in the train set. 
    

\xhdr{
Catboost Ranking model}\footnote{\url{https://catboost.ai/en/docs/concepts/loss-functions-ranking\#PairLogitPairwise}}  \cite{10.5555/3327757.3327770} using the ``PairLogitPairwise'' loss \cite{pmlr-v14-gulin11a}. 
This model takes groups (in our case, this corresponds to a round), and compares possible pairs of jokes within each group. Then it outputs an order on all group members. We used joke features using Catboost's text encoder. 

In addition, we tested several \emph{binary classification} algorithms. In those models, 
the input is one prompt and one punchline, and the task is to predict whether this punchline will be picked. To pick the winner of the round, we sort the punchlines by their predicted score and pick the top one. 
    
    \xhdr{Catboost gradient boosting tree classifier} Features include the Catboost built-in text encoder (seperately on the joke and prompt), pretrained deep learning embeddings (``all-MiniLM-L12-v2'', a 120M Sentence-Transformers model   \cite{reimers_sentence-bert_2019, DBLP:journals/corr/abs-2012-15828}), and the punchline's number of words and characters. 
        
    \xhdr{AutoML features + LightGBM classifier} We used {SparkBeyond,  a state-of-the-art machine learning (autoML) framework \cite{maor_system_2017}}. AutoML methods can help comprehensively and automatically find predictive signals in complex data \cite{cohen_icu_2021, feurer-neurips15a}. The system automatically extracts and ranks a wide range of features, including bag of words, interactions between text columns, pretrained embeddings, and semantic features (such as Wordnet and Wikipedia concepts).

    The top 300 features were used to train a LightGBM classification tree model  \cite{10.5555/3294996.3295074}.


Default hyperparameters were used for all models. Training time was a few minutes on a PC. Code (barring the autoML part) is provided in {\url{https://github.com/ddofer/CAH}}. 

We note we also experimented with Logistic regression and Random forest \cite{scikit-learn},  using embeddings from the pre-trained MiniLM-L12-v2 Sentence-Transformer model (``MiniLM DL'' in Table \ref{Table 1 - Predicting winning games}). 
Results were inferior (59 \& 56 AUC respectively), and this direction was abandoned.

\subsection{Results}
\label{sec:Predicting winning jokes}

\xhdr{Feature importance}
%
%
Feature importance  was ranked using marginalized mutual information gain. The top of the list was dominated by \emph{punchline} features (as opposed to features derived from prompt or combined joke text). In brief, \emph{punchline} features correlating to “dirty”, gross-out, sexually anatomical concepts appear at the top of the list (e.g., words belonging to the WordNet obscenity synset, relating to “pejorative terms for people”, drugs, sexual acts and male genitalia). Short punchlines (under 5 words) are also strongly preferred, 
with a 9\% higher win rate; we hypothesize that short phrases are easier to use in different contexts. (This is interesting, as there is evidence in humor literature supporting both short and long utterances \cite{kuipers2015good,ziv1984personality}).



Features of jokes with a \emph{low} success rate included a high, positive emotional sentiment score (e.g. “Having a wonderful time at the zoo"), involving edible things,  or relating to the WordNet hypernym of “cognition” (e.g., “body image”).

\xhdr{Prediction task}     
%
We computed ROC-AUC and top-1 accuracy (Acc@1, was the highest ranked card actually the winning card). A random baseline would get 10\% top-1 accuracy. 
Results are shown in Table \ref{Table 1 - Predicting winning games}.
We see that the Catboost classifier and autoML+LightGBM approaches perform about twice as good as random, but surprisingly, the simple punchline popularity baseline performs about the same (and even slightly better). 
First, we conclude that the problem is hard. We were surprised, given that the winning baseline did not even have access to the prompts.

Following this discovery, and the supporting evidence from our feature importance analysis,
and decided to perform
 ablation testing, training Catboost classifiers on punchline only, prompt only, and (combined) joke only.
Results (Table \ref{Table 1 - Predicting winning games}) support the conclusion that the classifier performance is primarily determined by the punchline card alone.

We note we trained one additional classification model (``Catboost-Meta"") that also had access to the order cards were displayed. This achieved the best results, indicating some potential {user behavioral bias (with cards in the center of the screen being preferentially picked).}

We believe that the experiments highlight the shortcoming of neural language models, leaving a lot of room for future work.

\begin{table}[t!]
\begin{tabular}{lll}
Model & Acc@1 & AUC \\
\hline \hline
Random baseline & 10 & 50 \\ \hline
Punchline Popularity & {\bf 20.7} & {\bf 64.4} \\
Joke Popularity & 15.6 & 55.8 \\ \hline
Catboost-Ranker & 18.7 & 61 \\
AutoML+LightGBM & 20.3 & 64.3 \\
{MiniLM DL} & 17.7 & 59.4  \\
Catboost-Classifier & 20.4 & 64.3 \\
Catboost-Meta & {\bf 21.1} & {\bf 64.7} \\ 
\hline \hline
Catboost-Punchline only & 20.4 & 64.1 \\
Catboost-Joke only & 19.3 & 63.2 \\
Catboost-Prompt only & 9.9 & 50 \\ 
\end{tabular}
\caption{Predicting winning games (\% accuracy and AUC). The trained models substantially outperform random and joke popularity baselines. 
Surprisingly, the punchline baseline outperformed most models. Catboost-Meta uses all classifier features as well as card display order, achieving the best results.     
The bottom of the table shows ablation for specific inputs only; interestingly, performance seems to be primarily determined by the punchline card. 
\label{Table 1 - Predicting winning games}}
\end{table}

%% file: 04task2.tex
Given the finding in Section \ref{task1}, we set out to evaluate if our models merely memorized funny cards, or whether they could {\bf generalize}{ to novel punchlines}.

\subsection{Methods}

 We constructed a new validation setup, partitioned at the level of punchlines as well as games (the later being necessary for top-k evaluation). In each iteration we split the data at the level of games, with 60 games (up to 600 jokes) in the test set. The remaining games are filtered so that punchlines are disjoint from the test set; these games become the train set. 
 This was repeated 500 times, for a total of 300,000 (not unique) unseen punchlines.

Our best performing method from Section \ref{methods1}, punchline popularity, is useless when facing an unseen punchline. Models needs to generalize to new punchlines, not just jokes, to predict funniness in this new setup, instead of merely memorizing the win-rate of known punchlines. Thus, we picked the next best model -- the Catboost classifier.


\subsection{Results}
\label{subsection:Novel Punchlines’ evaluation}


 The Catboost classifier 
 achieved ROC-AUC 56\% and a top-1 accuracy of 14.6\%. Top-2 accuracy was 26.8\% and top-3 -- 37.7\%.  Random guessing baseline is 10\% top-1, 20\% top-2, 30\% top-3. Thus, we conclude that the model's ability to generalize is modest, and some of the performance in our previous task can be explained by seeing the same cards in training.
%
We see this task as having the most potential for improvement in future work. 